\documentclass[journal]{IEEEtran}

\usepackage[utf8]{inputenc} 
\usepackage[T1]{fontenc}    
\usepackage[bookmarks=false]{hyperref}
\usepackage{url}            
\usepackage{booktabs}       
\usepackage{amsfonts}       
\usepackage{nicefrac}       
\usepackage{microtype}      
\usepackage{amsmath}
\usepackage{amssymb}
\usepackage{graphicx}
\usepackage{cleveref}

\usepackage{tablefootnote}

\usepackage[flushleft]{threeparttable}
\newcommand{\fixme}[2]{\ifx&#2&{\leavevmode\color{red}#1}\else{\leavevmode\color{red}FIXME\{}#1{\leavevmode\color{red}\}}\footnote{{\leavevmode\color{red}#2}}\PackageWarning{Fixme}{#1: #2}\fi}

\usepackage[hang]{subfigure}

\usepackage[usenames, dvipsnames]{color}

\usepackage{tikz}
\usetikzlibrary{arrows,%
                petri,%
                topaths,
                shapes}%
\usetikzlibrary{shapes.geometric}

\usepackage{tkz-berge}

\usepackage{pgfplots}

\usepgfplotslibrary{units,groupplots}
\pgfplotsset{compat=newest}

\definecolor{Paired-2}{RGB}{166,206,227}
\definecolor{Paired-1}{RGB}{31,120,180}
\definecolor{Paired-4}{RGB}{178,223,138}
\definecolor{Paired-3}{RGB}{51,160,44}
\definecolor{Paired-6}{RGB}{251,154,153}
\definecolor{Paired-5}{RGB}{227,26,28}
\definecolor{Paired-8}{RGB}{253,191,111}
\definecolor{Paired-7}{RGB}{255,127,0}
\definecolor{Paired-10}{RGB}{202,178,214}
\definecolor{Paired-9}{RGB}{106,61,154}
\definecolor{Paired-12}{RGB}{255,255,153}
\definecolor{Paired-11}{RGB}{177,89,40}

\def\BibTeX{{\rm B\kern-.05em{\sc i\kern-.025em b}\kern-.08em
    T\kern-.1667em\lower.7ex\hbox{E}\kern-.125emX}}
\begin{document}
\bstctlcite{IEEEexample:BSTcontrol}

\title{Learning to Skip Ineffectual Recurrent Computations in LSTMs
}

\author{Arash~Ardakani, Zhengyun~Ji, and Warren~J.~Gross\\
McGill~University, Montreal, Canada
\vspace{-20pt}}
\pagestyle{empty} 
\maketitle
\thispagestyle{empty}

\begin{abstract}
Long Short-Term Memory (LSTM) is a special class of recurrent neural
network, which has shown remarkable successes in processing sequential data.
The typical architecture of an LSTM involves a set of states and gates: the states
retain information over arbitrary time intervals and the gates regulate the
flow of information. Due to the recursive nature of LSTMs, they are
computationally intensive to deploy on edge devices with limited hardware
resources. To reduce the computational complexity of LSTMs, we first introduce
a method that learns to retain only the important information in the states by
pruning redundant information. We then show that our method can prune over
90\% of information in the states without incurring any accuracy degradation
over a set of temporal tasks. This observation suggests that a large fraction
of the recurrent computations are ineffectual and can be avoided to speed up
the process during the inference as they involve noncontributory
multiplications/accumulations with zero-valued states. Finally, we introduce a
custom hardware accelerator that can perform the recurrent computations using
both sparse and dense states. Experimental measurements show that performing
the computations using the sparse states speeds up the process and improves
energy efficiency by up to 5.2$\times$ when compared to
implementation results of the accelerator performing the computations using
dense states.
\end{abstract}

\vspace{-10pt}

\section{Introduction}
Convolutional neural networks (CNNs) have surpassed human-level accuracy in
different complex tasks that require learning hierarchical representation of
spatial data \cite{deeplearning}. CNN is a stack of multiple convolutional
layers followed by a few fully-connected layers \cite{conv}. The computational
complexity of CNNs is dominated by the multiply-accumulate operations of
convolutional layers as they perform high dimensional convolutions while the
majority of weights are usually found in fully-connected layers.\par

Recurrent neural networks (RNNs) have also shown remarkable success in
processing variable-length sequences. As a result, they have been adopted in
many applications performing temporal tasks such as language modeling
\cite{mikolov}, neural machine translation \cite{sutskever}, automatic speech
recognition \cite{grave_speech}, and image captioning \cite{vinyals}. Despite
the high prediction accuracy of RNNs over short-length sequences, they fail to
learn long-term dependencies due to the exploding gradient problem
\cite{pascanu}. Long Short-Term Memories (LSTM) \cite{lstm} is a popular class
of RNNs, that was introduced in literature to mitigate the above issue. Similar
to CNNs, LSTMs also suffer from high computational complexity due to the
recursive nature of LSTMs. \par

The recurrent transitions in LSTM are performed as 

\begin{equation}
\begin{pmatrix}
f_t \\
i_t \\
o_t \\
g_t
\end{pmatrix} =
\begin{pmatrix}
\sigma \\
\sigma \\
\sigma \\
\tanh
\end{pmatrix}
W_h h_{t-1} + W_x x_t + b,
\label{eq_lstm}
\end{equation}
\begin{equation}
c_t = f_t \odot c_{t-1} + i_t \odot g_t,
\label{eq_gates}
\vspace{-10pt}
\end{equation}
\begin{equation}
h_t = o_t \odot \tanh(c_t),
\label{eq_state}
\end{equation}
where $W_x \in \mathbb{R}^{d_x \times 4d_h}$, $W_h \in \mathbb{R}^{d_h \times
    4d_h}$ and $b \in \mathbb{R}^{4d_h}$ denote the recurrent weights and bias.
    The hidden states $h \in \mathbb{R}^{d_h}$ and $c \in \mathbb{R}^{d_h}$
    retain the temporal state of the network. The gates $f_t$, $i_t$, $o_t$ and
    $g_t$ regulate the update of LSTM parameters. The logistic sigmoid function
    and element-wise product are denoted as $\sigma$ and $\odot$, respectively.
    Eq. (\ref{eq_lstm}) involves the multiplication between the weight matrices
    (i.e., $W_h$ and $W_x$) and the vectors $x_t$ and $h_{t-1}$ while Eq.
    (\ref{eq_gates}) and Eq. ({\ref{eq_state}) perform only element-wise
        multiplications. Therefore, the recurrent computations of LSTM are
        dominated by the vector-matrix multiplications (i.e., $W_h h_{t-1} +
        W_x x_t$). It is worth mentioning that LSTM architectures are usually
        built on high dimensional input (i.e., $x_t$) or state (i.e.,
        $h_{t-1}$), increasing the number of operations performed in Eq.
        (\ref{eq_lstm}). Moreover, the recurrent computations of LSTM are
        executed sequentially, as the input at time $t$ depends on the output at
        time $t-1$. Therefore, the high computational complexity of LSTMs makes
        them difficult to deploy on portable devices requiring real-time
        processes at the edge.\par

To reduce the computational complexity of CNNs, recent convolutional
accelerators have exploited pruning \cite{EIE}, binarization \cite{iscas-binary},
and intrinsic sparsity among activations \cite{cnvlutin}, improving both
processing time and energy efficiency. Exploiting pruning to speed up the
recurrent computations of LSTMs has been also studied in \cite{ese}. However,
no practical accelerator for recurrent computations was introduced in
literature since the binarization algorithms developed for LSTMs are limited to
certain temporal tasks such as word language modelling, and do not generalize
well over different temporal tasks \cite{alternating}. The convolutional
accelerator introduced in \cite{cnvlutin} exploits the intrinsic sparsity among activations
incurred by the rectified linear unit (ReLU) in order to avoid ineffectual
multiplications/accumulations with zero-valued activations. Despite its
remarkable performance, this technique has not been exploited in
architectures accelerating recurrent computations yet, since LSTMs use sigmoid
and tanh functions as their non-linear functions.\par

Motivated by the above statements, this paper aims to first introduce a method
that learns only the important information stored in to the state $h_{t-1}$ as
this state contributes the most in the recurrent computations (see Eq.
(\ref{eq_lstm})). Note that the input vector $x_t$ is usually one-hot encoded
except for certain tasks with high dimensional vocabulary sizes such as word
language modelling. In this case, the embedding layer is used for these
tasks to reduce the input dimension \cite{mikolov}. In either cases, $d_h$ is
usually greater than $d_x$. After introducing the learning method, we show that
a large fraction of information coming from previous time steps are unimportant
and can be dropped. More precisely, the proposed learning algorithm can prune
over 90\% of values in the state vector $h_{t-1}$ over different temporal tasks
such as character language modeling, word language modeling and sequential
image classification without incurring any performance degradation. We then
introduce a custom hardware accelerator that can perform the recurrent
computations on both sparse and dense representations of the state vector
$h_{t-1}$. Finally, we show that performing the computations on the sparse
state vector results in up to 5.2$\times$ speedup and energy improvement over
the dense representation when both running on the proposed accelerator. To the
best of our knowledge, this work is the first attempt to skip the ineffectual
recurrent computations incurred by pruning the hidden state $h_{t-1}$.\par

\section{Learning Ineffectual Recurrent Computations}
During the past few years, it has been widely observed that deep neural
networks contain numerous redundant operations that can be removed without
incurring any accuracy degradation \cite{ese}. In \cite{ese,EIE}, it was shown
that exploiting pruned weights can speed up the computations of LSTMs and CNNs
by skipping the multiplications/accumulations with zero-valued weights. In
\cite{cnvlutin}, the intrinsic sparsity among the activations in CNNs was
exploited to speed up the convolutional computations. In fact, CNNs commonly
use the ReLU as their non-linear function, which dynamically clamps all
negatively-valued activations to zero. Unlike CNNs, LSTMs use sigmoid and tanh
units as their non-linear function. As a result, in order to use the latest
technique to speed up the recurrent computations, we first need a training
method to learn ineffectual computations among the activations (i.e., the state
$h_{t-1}$) in LSTMs.\par

\subsection{Pruning Method} \label{pruning-method}
The total number of operations required to perform Eq. (\ref{eq_lstm}) is equal
to $2 \times (d_x \times 4 d_h + d_h \times 4 d_h) + 4 d_h$ when considering
each multiply-accumulate as two operations. In temporal tasks such as
character-level language modeling, where the input vector $x_t$ is one-hot encoded,
its size (i.e., $d_x$) is limited to a few hundreds. In this case, the
vector-matrix multiplication of $W_x x_t$ is implemented as a look-up table and
its number of operations is equal to $4 d_h$, similar to the number of
operations required for biases. Eq. (\ref{eq_gates}) and Eq. (\ref{eq_state})
also require $3d_h$ and $d_h$ operations, respectively. Therefore, the
computational core of LSTMs is dominated by the matrix-vector multiplication of
$W_h h_{t-1} + W_x x_t$.\par

\begin{figure}[!t]
\centering
\includegraphics[scale = 0.45]{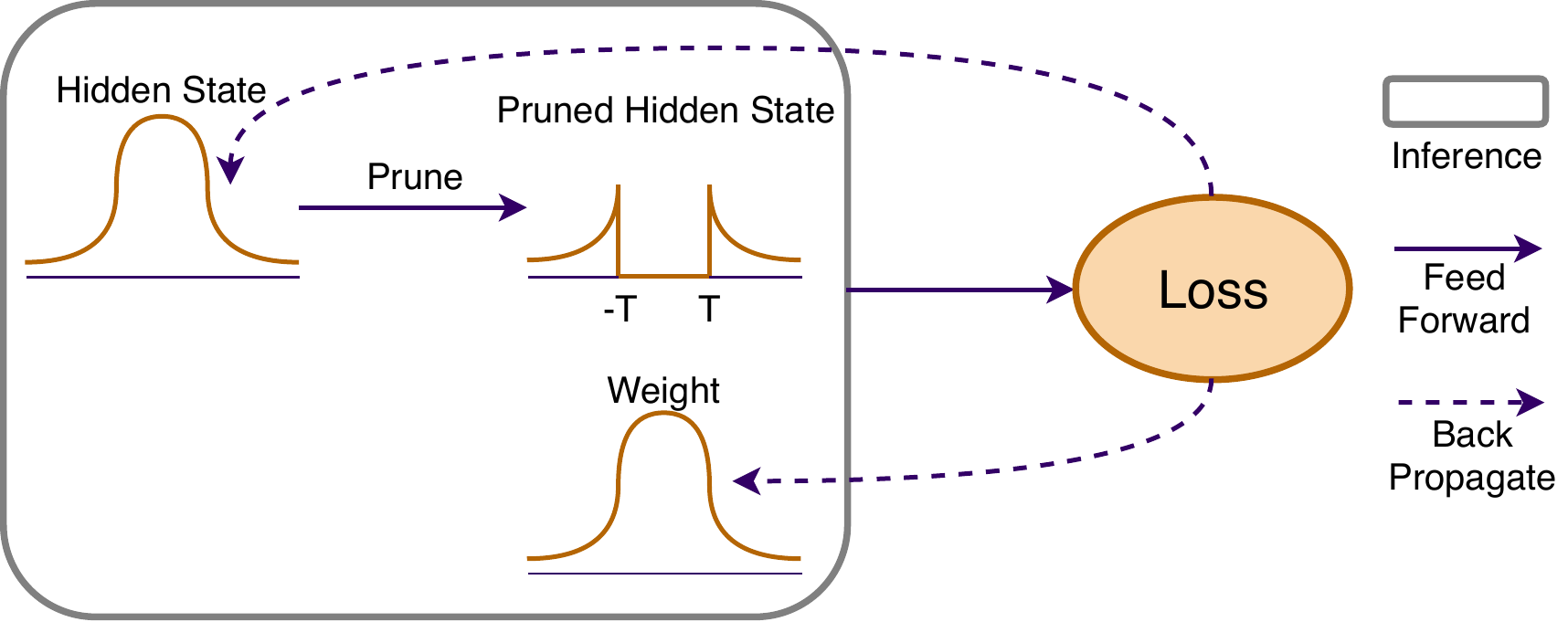}
\caption{The pruning procedure.}
\label{training_algorithm}
\vspace{-10pt}
\end{figure}

In order to reduce the computational complexity of LSTMs, we train LSTMs such
that they learn to retain only the important information in their hidden states
$h_{t-1}$. In fact, we train LSTMs with a sparsity constraint: all values below
the threshold $T$ are pruned away from the network, as shown in Fig.
\ref{training_algorithm}. During the training phase, we prune the state vector
$h_{t-1}$ when performing the feed-forward computations while its dense
representation is used during the update process of the LSTM parameters. In
fact, the feed-forward computations of LSTMs remains the same as the
conventional computations except for Eq. (\ref{eq_lstm}) which can be
mathematically formulated as

\begin{equation}
\begin{pmatrix}
f_t \\
i_t \\
o_t \\
g_t
\end{pmatrix} =
\begin{pmatrix}
\sigma \\
\sigma \\
\sigma \\
\tanh
\end{pmatrix}
W_h h^p_{t-1} + W_x x_t + b,
\label{eq_lstm_pruned}
\end{equation}
where $h^p_{t-1}$ denotes the sparse state vector and is obtained by
\begin{equation}
h^p_{t-1} =
  \begin{cases}
    0       & \quad \text{if } \lvert h_{t-1}\rvert < T \\
    h_{t-1}  & \quad \text{if } \lvert h_{t-1}\rvert \geq T
  \end{cases}.
\end{equation}

Maintaining the dense representation of the state vector during the update
process allows the state values initially lied within the threshold to be
updated. This technique was first introduced in \cite{BiCon} to binarize the
weights, while we use it to prune the state vector in our work. During the
update process, we need to compute the gradient on the hidden state $h_0$.
Due to the discontinuity of the rectangular function used to obtain $h^p_{t-1}$
at the threshold value, we approximate the derivatives on the hidden state $h_0$
by computing
\begin{equation}
\dfrac{\partial L}{\partial h_0} \approx \dfrac{\partial L}{\partial h^p_0}.
\end{equation}
\par
\subsection{Experimental Results}\label{Experimental-Results}
In this section, we evaluate the performance of the proposed training algorithm
that prunes away the noncontributory information of the hidden state vector
$h_{t-1}$ to the prediction accuracy on different temporal tasks: classification
of hand-written digits on sequential MNIST \cite{pmnist} and  both
character-level and word-level language modeling tasks on Penn Treebank corpus
\cite{PTB}. Since the pruning threshold is empirical, we report the prediction
accuracy of the above tasks for different sparsity degrees while using an 8-bit
quantization for all weights and input/hidden vectors. Of course, we cannot
always retain the prediction accuracy at the same level of the dense model for
any sparsity degree, as pruning too much hurts the prediction accuracy.\par

\subsubsection{Character-Level Language Modeling}
The main goal of character-level language modeling is to predict the next
character. For this task, the performance is measured by the bits per character
(BPC) metric, where a low BPC is desirable.
For this task, we use an LSTM layer with 1000 units (i.e., $d_h$) followed by a
classifier. The model is then trained on the Penn Treebank dataset with a sequence
length of 100. We minimize
the cross entropy loss using stochastic gradient descent on mini-batches of size
64. The update rule is determined by ADAM with a learning rate of 0.002.\par

For the data preparation, we use the same configuration as \cite{mikolov}: we
split this dataset into 5017k, 393k and 442k training, validation and test
characters, respectively. Penn Treebank corpus has a vocabulary size of 50. For
the character-level language modeling task, the input vector is one-hot
encoded.\par

Fig. \ref{BPC} reports the prediction accuracy on the test set for different
sparsity degrees in terms of BPC. The experimental results show that over 90\%
of information in the hidden states can be pruned away without incurring any
degradation in the prediction accuracy. We can also see some improvements in
the prediction accuracy in the pruned models. In fact, the pruning algorithm
acts as a form of regularization that prevents the model from over-fitting. For
implementation purposes, we focus on the model with the sparsity degree of 97\%
(i.e., the sweet spot), which results in no accuracy degradation compared to
the dense model.

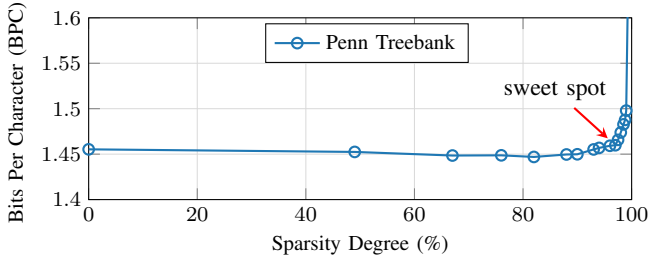
\begin{figure}[!t]
\footnotesize	

\centering
\scalebox{1}{
\begin{tikzpicture}

\begin{groupplot}[group style={group size=1 by 3},height=4cm, width=8.8cm]
    \nextgroupplot[every axis plot/.append style={thick}, 
                   grid=both, 
                   grid style={line width=.1pt, draw=gray!30}, 
                   ylabel style={align=center}, 
                   ylabel={Bits Per Character (BPC)}, 
                   xlabel = {Sparsity Degree (\%)}, 
                   legend style ={ at={(0.7,0.97)}}, 
                   ymin=1.4, 
                   ymax=1.6, 
                   xmin = 0, 
                   xmax = 100]

\addplot[color=Paired-1, mark = o]
  coordinates{
    (0,1.455287)
    (49,1.452441)
    (67,1.448481)
    (76,1.448702)
    (82,1.446922)
    (88,1.449627)
    (90,1.449845)
    (93,1.455179)
    (94,1.456862)
    (96,1.459201)
    (97,1.459798)
    (97.5,1.465730)
    (98,1.473847)
    (98.5,1.482738)
    (98.8,1.487671)
    (99,1.4979)
    (100,1.913647)
};
\addlegendentry{Penn Treebank}

\node[anchor=west, thick, ] (source) at (axis cs:75,1.52){\small sweet spot};
       \node (destination) at (axis cs:97,1.459798){};
       \draw[->,red, thick,-stealth](source)--(destination);

\end{groupplot}
\end{tikzpicture}
\label{LC}}

\caption{Prediction accuracy of character-level language modeling on the test set of the Penn Treebank corpus for different sparsity degrees over a sequence length of 100.}
    \label{BPC}
    \vspace{-10pt}
\end{figure}

\begin{figure}[!t]
\footnotesize	
\centering
\scalebox{1}{
\begin{tikzpicture}

\begin{groupplot}[group style={group size=1 by 3},height=4cm, width=8.8cm]
    \nextgroupplot[every axis plot/.append style={thick}, grid=both, grid style={line width=.1pt, draw=gray!30}, legend style ={ at={(0.7,0.97)}}, ymin=85, ymax=95, xmin = 0, xmax = 100, ,xlabel={Sparsity Degree (\%)}, ylabel style={align=center}, ylabel={Perplexity Per\\Word (PPW)} ]

\addplot[color=Paired-1, mark = o]
  coordinates{
    (0,87.937539)
    (67,86.432628)
    (72,86.031119)
    (80,86.055841)
    (82,86.429401)
    (86,86.429401)
    (90,87.486678)
    (92,87.514628)
    (93,87.978131)
    (95,91.467540)
    (96,91.410461)
    (98,93.181545)
    (100,97.113767)
};
\addlegendentry{Penn Treebank}

\node[anchor=west, thick, ] (source) at (axis cs:59,91){\small sweet spot};
       \node (destination) at (axis cs:93,87.978131){};
       \draw[->,red, thick,-stealth](source)--(destination);

\end{groupplot}
\end{tikzpicture}
}

\caption{Prediction accuracy of word-level language modeling on the test set of Penn Treebank for different sparsity degrees over a sequence length of 35.}
    \label{PPW}
    \vspace{-10pt}
\end{figure}
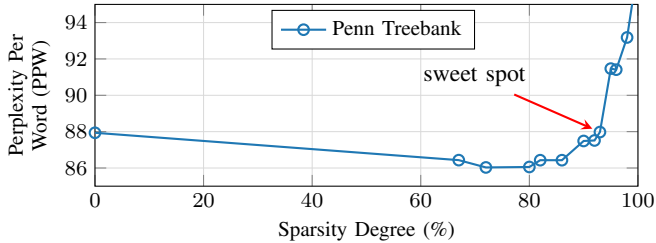

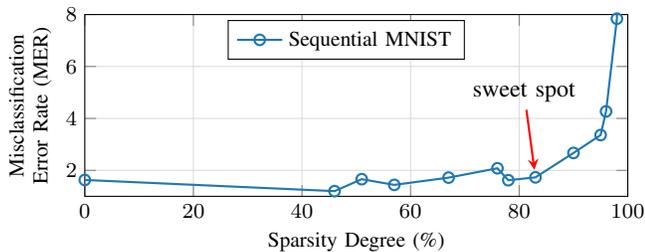
\begin{figure}[!t]
\footnotesize	
\centering
\scalebox{1}{
\begin{tikzpicture}

\begin{groupplot}[group style={group size=1 by 3},height=4cm, width=8.8cm]
    \nextgroupplot[every axis plot/.append style={thick}, grid=both, grid style={line width=.1pt, draw=gray!30}, legend style ={ at={(0.7,0.97)}}, ymin=1, ymax=8, xmin = 0, xmax = 100, ,xlabel={Sparsity Degree (\%)}, ylabel style={align=center}, ylabel=Misclassification\\Error Rate (MER) ]

\addplot[color=Paired-1, mark = o]
  coordinates{
    (0,1.63)
    (46,1.2)
    (51,1.66)
    (57,1.44)
    (67,1.72)
    (76,2.08)
    (78,1.62)
    (83,1.73)
    (90,2.67)
    (95,3.36)
    (96,4.27)
    (98,7.84)
    (100,53.2)
};
\addlegendentry{Sequential MNIST}

\node[anchor=west, thick, ] (source) at (axis cs:70,5){\small sweet spot};
       \node (destination) at (axis cs:83,1.73){};
       \draw[->,red, thick,-stealth](source)--(destination);

\end{groupplot}
\end{tikzpicture}
}

\caption{Misclassification error rate of image classification task on the test set of MNIST.}
    \label{MER}
    \vspace{-10pt}
\end{figure}

\subsubsection{Word-Level Language Modeling}
For the word-level task, we use Penn Treebank corpus with a 10K vocabulary
size. Similar to \cite{mikolov}, we split the dataset into 929K training, 73K
validation and 82K test tokens. We use one LSTM layer of size 300 followed by a
classifier layer to predict the next word. The performance is evaluated on
perpelexity per word (PPW). Similar to character-level tasks, the models with
lower PPW are better. We also use an embedding layer of size 300 to reduce the
dimension of the input vector \cite{mikolov}. Therefore, the input vector $x_t$
contains real values as opposed to the character-level modeling task in which
the input vector is one-hot encoded.\par

We train the model with the word sequence of 35 while applying the dropout
probability of 0.5 on the non-recurrent connections similar to \cite{zaremba}.
We train the network using the learning rate of 1 and the learning decay factor
of 1.2. We also clip the norm of the gradients to 5. Fig. \ref{PPW} shows the
predication accuracy of the pruned models performing the word-level task on the
test set of Penn Treebank. From the experimental results, we observe that over
90\% of information in the hidden state can be pruned away without affecting
the prediction accuracy.\par

\begin{figure*}[!t]
    \centering
    \subfigure[]{
        \includegraphics[scale = 0.6]{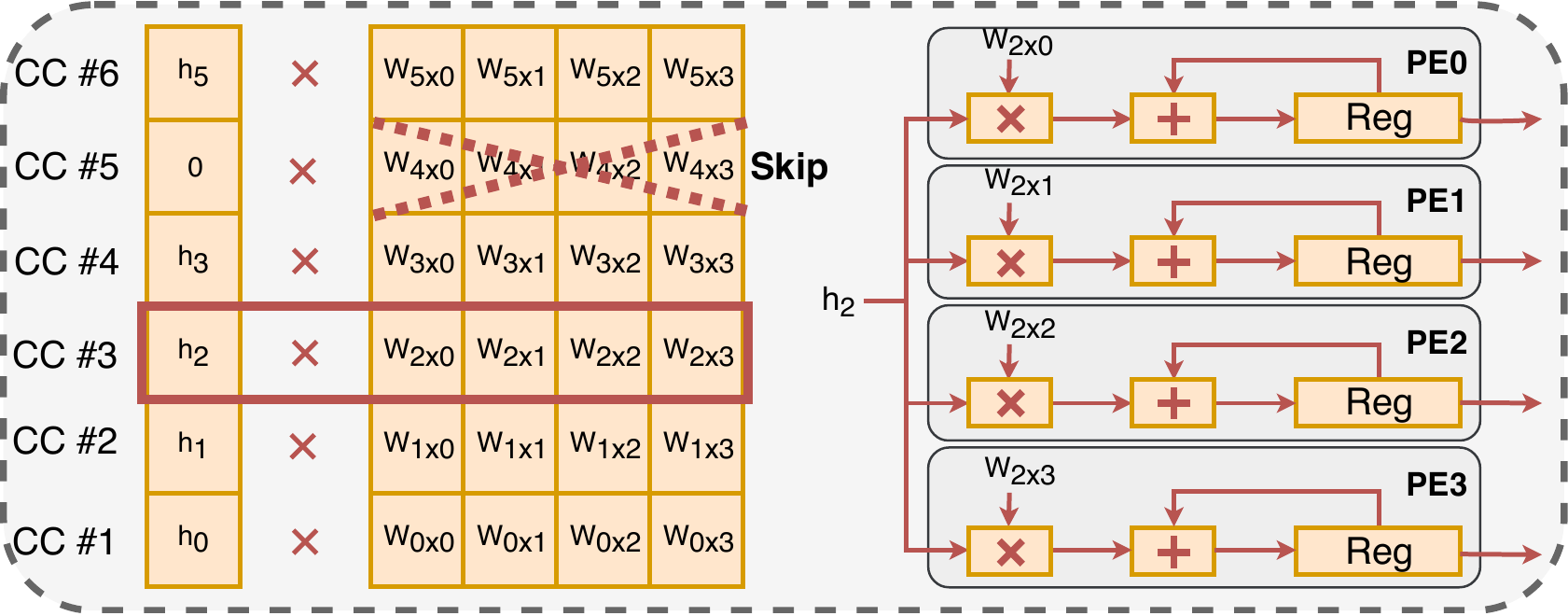}
        \label{semi-imp-1}
    }
    \subfigure[]{
        \includegraphics[scale = 0.6]{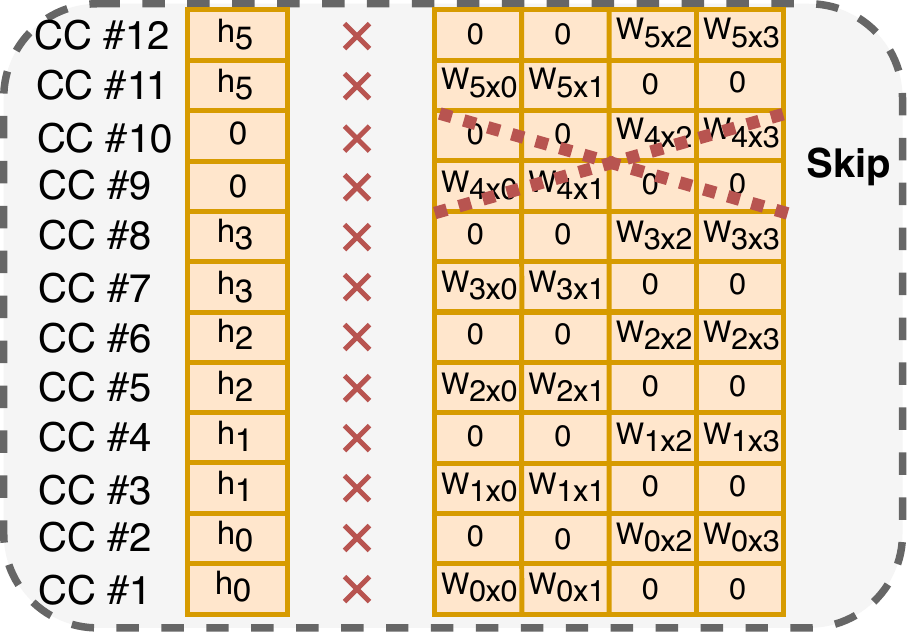}
        \label{semi-imp-2}
    }
    \subfigure[]{
        \includegraphics[scale = 0.57]{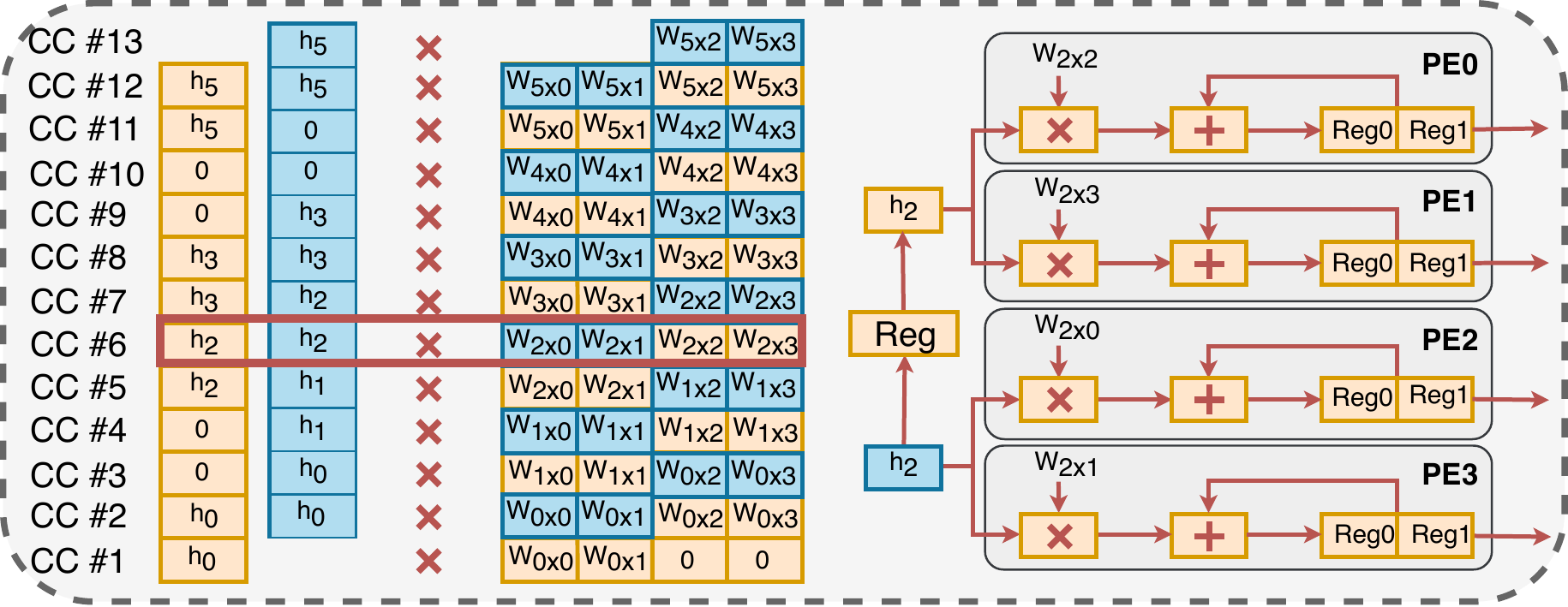}
        \label{semi-imp-3}
    }
    \subfigure[]{
        \includegraphics[scale = 0.58]{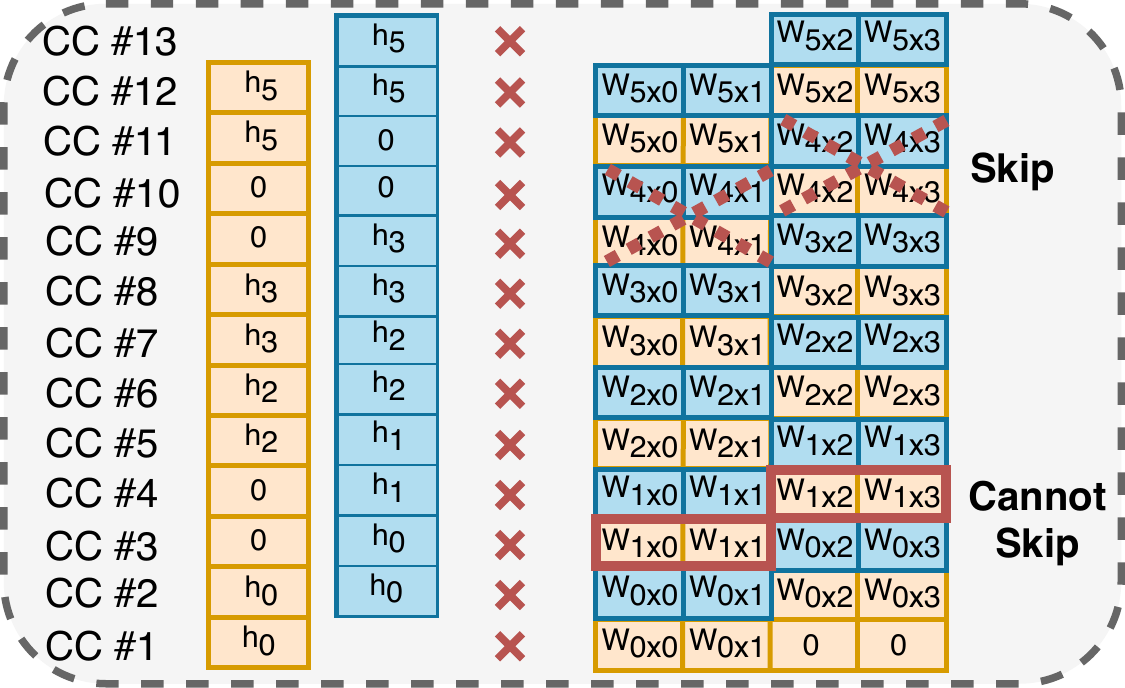}
        \label{semi-imp-4}
    }
    \caption{A vector-matrix multiplication: (a) under an unlimited data bandwidth with a batch size of 1, (b) under a limited data bandwidth with a batch size of 1 and (c) under a limited data bandwidth with a batch size of 2. (d) Skipping is allowed only when both batches contain zero-valued inputs at the same position. }
    \label{mult}
\end{figure*}

\subsubsection{Image Classification}
We perform an image classification task on MNIST dataset containing 60000
gray-scale images (50000 for training and 10000 for testing), falling into 10
classes. For this task, we process the pixels in scanline order: each image
pixel is processed at each time step similar to \cite{pmnist}. To this end, we
use an LSTM layer of size 100 and a softmax classifier layer. We also adopt
ADAM step rule with the learning rate of 0.001. The misclassification error
rate of the models with pruned states is illustrated in Fig. \ref{MER}. Similar
to the language modeling tasks, the MNIST classification task can be performed
with the models exploiting over 80\% pruned hidden state without affecting the
misclassification error rate.\par

\section{Hardware Implementation}
\vspace{-3pt}
So far, we have shown that over 90\% of the hidden vector
$h_{t-1}$ can be pruned away without any accuracy degradation over a set of
temporal tasks. This observation suggests that over 10$\times$ speedup is
expected when performing the matrix-vector multiplication $W_h h_{t-1}$.
However, exploiting the sparse hidden vector in specialized hardware to speed
up the recurrent computations is nontrivial and introduces new challenges. The
first challenge is to design a dataflow to efficiently perform the recurrent
computations, as the LSTM network involves two different types of computations:
the matrix-vector multiplication (i.e., Eq.(\ref{eq_lstm})) and the Hadamard
product (i.e., Eq. (\ref{eq_gates}) and Eq. (\ref{eq_state})). Secondly, the
bandwidth of the off-chip memory is limited, making the fully exploitation of
the parallelism difficult. Finally, a customized decoding scheme is required to
denote the indices to the unpruned information. In this section, we propose a
dataflow for recurrent computations and its architecture, which efficiently
exploit the pruned hidden vector to skip the ineffectual computations under a
bandwidth limited interface with the off-chip memory. \par

\subsection{Recurrent Computation Dataflow} \label{dataflow}
As discussed in Section \ref{pruning-method}, the main computational core of
LSTM involves two vector-matrix multiplications. A typical approach to
implement a vector-matrix multiplications is a semi-parallel implementation, in
which a certain number of processing elements (PEs) are instantiated in parallel
to perform multiply-accumulate operations. In this approach, all the PEs share
the same input at each clock cycle and the computation is performed serially.
Fig. \ref{semi-imp-1} illustrates an example performing a vector-matrix
multiplication on an input vector of size 6 and a weight matrix of size
4$\times$6. In this example, 4 PEs are required in parallel to perform the
computations serially within 6 clock cycles. Clearly, the computation of the
zero-valued elements of the input vector can be skipped to speed up the
process. However, in a larger scale, this approach requires high memory
bandwidth to provide weights for the PEs in parallel. Let us consider that the
memory bandwidth provides only a single input element and two weight values
at each clock cycle as an example. In this case, 12 clock cycles are required
to perform the computations when using 4 PEs in parallel.
In fact, the latency is doubled for the bandwidth limited scenario while the
utilization factor of PEs are reduced to 50\% as shown in Fig.
\ref{semi-imp-2}.\par

To increase the throughput and utilization factor of the bandwidth limited
scenario, we use a higher batch size. In this case, we need to use more memory
elements to store the partial values for each batch. For instance, we use a
batch size of 2 to fill the empty pipeline stages in Fig. \ref{semi-imp-2}.
More precisely, the first input element of the first batch and the first two
weights for the first two PEs are read. The weights for the PEs are then stored
into the registers to be used for the first input element of the next batch.
Meanwhile, the first two PEs perform the multiply-accumulate operations between
the weights and the first input element of the first batch. The partial
result is stored into the scratch memory cells. In the second clock cycle, the
first input element of the second batch along with the weights for the second
two PEs are read from the off-chip memory. Meanwhile, the first element of the
first batch is passed to the second two PEs through the pipeline stages.
Therefore, the computations of the second two PEs are performed with the first
element of the first batch and the computations of the first two PEs with the
first element of the second batch. After 2 clock cycles for this example, all
the PEs process the multiply-accumulate operations in parallel and the pipeline
stages are full, resulting in a high utilization factor as shown in Fig.
\ref{semi-imp-3}.\par

While using a higher batch size results in a higher utilization factor and
throughput, each positional element of all the batches has to be zero in order
to skip the computations of its corresponding clock cycles. Let us consider the
previous example again. If the second input element of the first batch is zero
while the second input element of the second batch is non-zero, we cannot skip
its corresponding clock cycles since both batches share the same weights as
depicted in Fig. \ref{semi-imp-4}. Therefore, we can only skip those
computations in which all the input elements of the all batches are zero (see
Fig. \ref{semi-imp-4}).\par

\subsection{Zero-State-Skipping Accelerator}
Since this paper targets portable devices at the edge with limited hardware
resources and memory bandwidth, we use the aforementioned scheme to exploit the
sparse hidden vector in order to accelerator the recurrent computations. Fig. \ref{architecture}
shows the detailed architecture of the proposed accelerator for LSTM networks.
To this end, we exploit four tiles: each tile contains 48 PEs performing
multiply-accumulate operations. In fact, each tile is responsible to obtain the
values of one gate by computing Eq. (\ref{eq_lstm_pruned}). Therefore, the
first three tiles and the fourth tile are equipped with sigmoid and tanh
functions, respectively. We also adopt LPDDR4 as the off-chip memory, which
easily provides a bandwidth of 51.2 Gbps \cite{DRAM}. More precisely, this
data bandwidth provides 24 8-bit weights and a single 8-bit input element for
24 PEs at a nominal frequency of 200 MHz. As a result, we associate each PE with
a 16$\times$12-bit scratch memory to store the partial products for 16 batch
sizes. For the Hadamard products of Eq. (\ref{eq_gates}), the first tile
performs the element-wise product of the first term (i.e., $f_t \odot c_{t-1}$)
by reading the values of $c_{t-1}$ from the off-chip memory. Meanwhile, the
computations for the second term ($i_t \odot g_t$) is performed in the second
tile. Then, the results of the both terms are passed to the fourth tile to
perform the addition and obtain tanh($c_t$). It is worth mentioning that all
the PEs can take inputs from either the off-chip memory or the scratch memory
of the same tile or others through the global and local routers
(see Fig. \ref{architecture}). Finally, the computations of Eq. (\ref{eq_state})
are performed in the third tile to obtain $h_t$. The obtained results are then passed
to an encoder that keeps track of zero-valued elements using a counter.
More precisely, the encoder counts up if the current input value of all the
batches is zero. Afterwards, the obtained offset is stored along with the hidden
state vector into the off-chip memory. During the recurrent computations of the
next time step, the offset is only used to read the weights that correspond
to the non-zero values. Therefore, no decoder is required in this scheme.
The weight/input registers in Fig. \ref{architecture} are also used to provide
the pipeline stages for the batches.\par

\begin{figure}[!t]
\centering
\includegraphics[scale = 0.5]{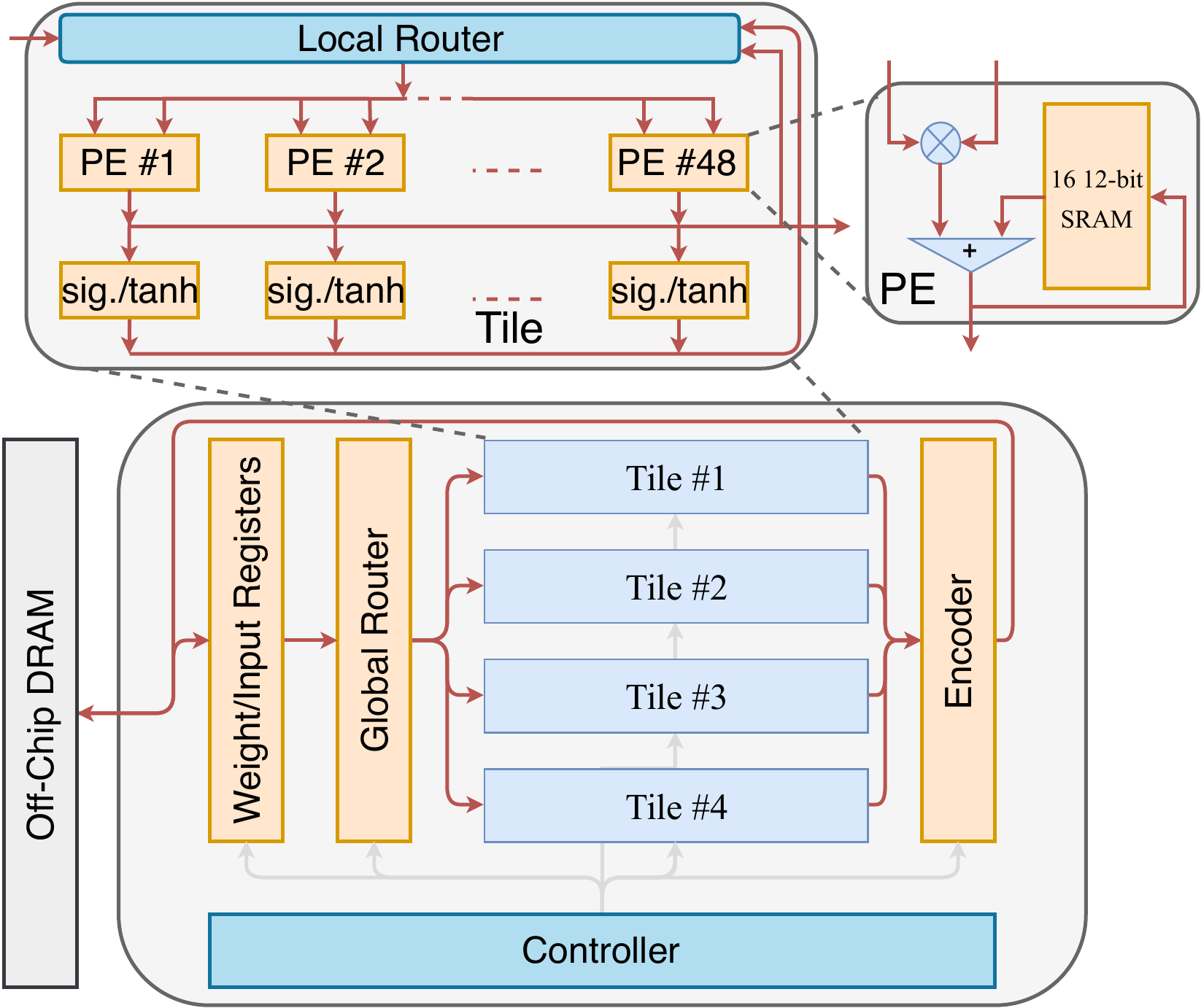}
\caption{The proposed LSTM accelerator.}
\label{architecture}
\end{figure}

\begin{figure}[!t]
\centering
\small
\scalebox{0.8}{
\begin{tikzpicture}
  \centering
  \begin{axis}[
        ybar, axis on top,
        height=4cm, width=9cm,
        bar width=0.4cm,
        ymajorgrids, tick align=inside,
        major grid style={draw =white},
        enlarge y limits={value=.1,upper},
        ymin=0, ymax=140,
        axis x line*=bottom,
        axis y line*=left,
        y axis line style={opacity=0},
        tickwidth=0pt,
        enlarge x limits={abs = 1.5cm},
        x = 3cm,
        legend style={
            at={(0.5,-0.2)},
            anchor=north,
            legend columns=-1,
            /tikz/every even column/.append style={column sep=0.5cm}
        },
        ylabel={Sparsity Degree (\%)},
        symbolic x coords={
         1 batch, 8 batches,16 batches  },
       xtick=data,
       nodes near coords=\rotatebox{90}{
        \pgfmathprintnumber[precision=1]{\pgfplotspointmeta}
       }
    ]
    \addplot [draw=none, fill=blue!40] coordinates {
      (1 batch, 97)
      (8 batches, 81)
      (16 batches,66) };
   \addplot [draw=none,fill=red!40] coordinates {
      (1 batch, 93)
      (8 batches, 63)
      (16 batches,41) };
   \addplot [draw=none,fill=teal!60] coordinates {
      (1 batch, 83)
      (8 batches, 55)
      (16 batches,43) };

    \legend{PTB-Char, PTB-Word,
          MNIST}
  \end{axis}
  \end{tikzpicture}}
\caption{Sparsity degree of the hidden state vector over different batch sizes.}
\label{SD}
\end{figure}
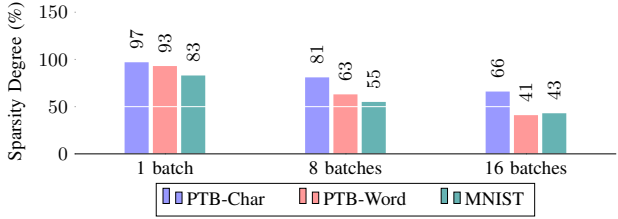

\subsection{Methodology}
We implemented the proposed accelerator in Verilog and synthesized via Cadence
Genus Synthesis Solution using TSMC 65nm GP CMOS technology. We also used
Artisan dual-port memory compiler to implement the scratch memories for PEs.
The proposed accelerator occupies a silicon area of 1.1 mm$^2$ and yields a peak
performance of 76.8 Gops and performance efficiency of 925.3 Gops/W over dense models at a nominal
frequency of 200 MHz. For evaluation purposed, we perform the recurrent
computations on the proposed accelerator at its nominal frequency over a set of
temporal tasks when using both dense and sparse representations for the hidden
state vector $h_{t-1}$.

\begin{figure}[!t]
\centering
\small
\scalebox{0.8}{
\begin{tikzpicture}
  \centering
  \begin{axis}[
        ybar, axis on top,
        height=4.5cm, width=9cm,
        bar width=0.3cm,
        ymajorgrids, tick align=inside,
        major grid style={draw =white},
        enlarge y limits={value=.1,upper},
        ymin=0, ymax=400,
        axis x line*=bottom,
        axis y line*=left,
        y axis line style={opacity=0},
        tickwidth=0pt,
        enlarge x limits={abs = 1.5cm},
        x = 3cm,
        legend style={
            at={(0.5,-0.2)},
            anchor=north,
            legend columns=3,
            /tikz/every even column/.append style={column sep=0.5cm}
        },
        ylabel={Performance (GOPS)},
        symbolic x coords={
         1 batch, 8 batches,16 batches  },
       xtick=data,
       nodes near coords=\rotatebox{90}{
        \pgfmathprintnumber[precision=1]{\pgfplotspointmeta}
       }
    ]
    \addplot [draw=none, fill=blue!40] coordinates {
      (1 batch, 9.6)
      (8 batches, 76.4)
      (16 batches,76.4) };
   \addplot [draw=none,fill=red!40] coordinates {
      (1 batch, 314.7)
      (8 batches, 395.5)
      (16 batches,223) };
   \addplot [draw=none,fill=teal!60] coordinates {
      (1 batch, 9.6)
      (8 batches, 76.2)
      (16 batches,76.2) };
   \addplot [draw=none,fill=gray!60] coordinates {
      (1 batch, 17.9)
      (8 batches, 110.8)
      (16 batches,95.6) };

   \addplot [draw=none,fill=green!50] coordinates {
      (1 batch, 9.6)
      (8 batches, 74.3)
      (16 batches,74.3) };
   \addplot [draw=none,fill=cyan!60] coordinates {
      (1 batch, 50.5)
      (8 batches, 154.3)
      (16 batches,124.9) };

    \legend{PTB-Char(Dense), PTB-Char(Sparse), PTB-Word(Dense), PTB-Word(Sparse),
          MNIST(Dense), MNIST(Sparse)}
  \end{axis}
  \end{tikzpicture}}
\caption{Performance of the proposed accelerator over dense and sparse models.}
\label{performance}
\end{figure}
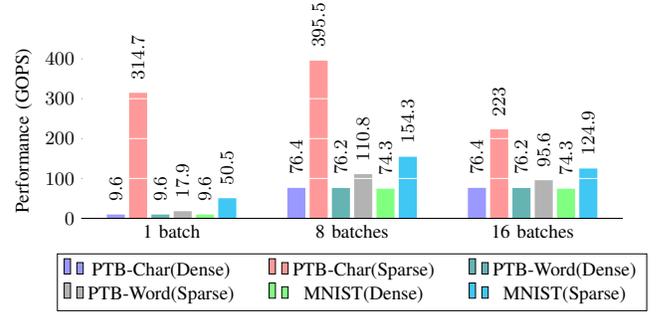

\begin{figure}[!t]
\centering
\small
\scalebox{0.8}{
\begin{tikzpicture}
  \centering
  \begin{axis}[
        ybar, axis on top,
        height=4.5cm, width=9cm,
        bar width=0.3cm,
        ymajorgrids, tick align=inside,
        major grid style={draw =white},
        enlarge y limits={value=.1,upper},
        ymin=0, ymax=4800,
        axis x line*=bottom,
        axis y line*=left,
        y axis line style={opacity=0},
        tickwidth=0pt,
        enlarge x limits={abs = 1.5cm},
        x = 3cm,
        legend style={
            at={(0.5,-0.2)},
            anchor=north,
            legend columns=3,
            /tikz/every even column/.append style={column sep=0.5cm}
        },
        ylabel={Energy Efficiency (GOPS/W)},
        symbolic x coords={
         1 batch, 8 batches,16 batches  },
       xtick=data,
       nodes near coords=\rotatebox{90}{
        \pgfmathprintnumber[precision=1]{\pgfplotspointmeta}
       }
    ]
    \addplot [draw=none, fill=blue!40] coordinates {
      (1 batch, 115.7)
      (8 batches, 920.5)
      (16 batches,920.5) };
   \addplot [draw=none,fill=red!40] coordinates {
      (1 batch, 3791.6)
      (8 batches, 4765.1)
      (16 batches,2686.7) };
   \addplot [draw=none,fill=teal!60] coordinates {
      (1 batch, 115.7)
      (8 batches, 918.1)
      (16 batches,918.1) };
   \addplot [draw=none,fill=gray!60] coordinates {
      (1 batch, 215.7)
      (8 batches, 1335)
      (16 batches,1151.8) };

   \addplot [draw=none,fill=green!50] coordinates {
      (1 batch, 115.7)
      (8 batches, 895.2)
      (16 batches,895.2) };
   \addplot [draw=none,fill=cyan!60] coordinates {
      (1 batch, 608.4)
      (8 batches, 1859)
      (16 batches,1504.8) };

    \legend{PTB-Char(Dense), PTB-Char(Sparse), PTB-Word(Dense),\newline PTB-Word(Sparse),
          MNIST(Dense), MNIST(Sparse)}
  \end{axis}
  \end{tikzpicture}}
\caption{Energy efficiency of the proposed accelerator over dense and sparse models.}
\label{energy}
\end{figure}
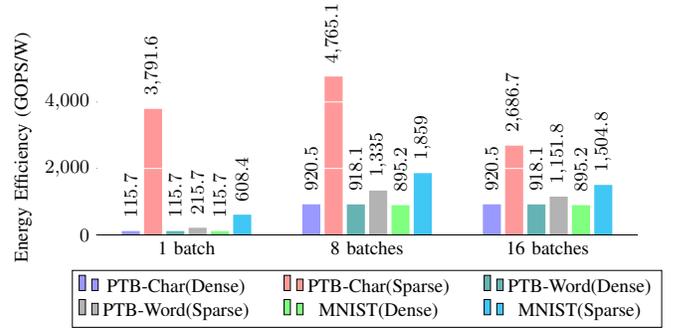

\subsection{Implementation Results}
In Section \ref{Experimental-Results}, we showed that the accuracy curves of
different temporal tasks for different sparsity degree when using a batch size
of one. However, due to the limited bandwidth of the off-chip memory, we use higher
batch sizes to increase the utilization factor of PEs and improve the
throughput of the accelerator (see Section \ref{dataflow}). To exploit the
sparsity among information of the hidden state in order to speed up the
recurrent computations, all the elements of all the batches must be zero. For
instance, all the third elements of all the batches must be zero to skip their
ineffectual computations. As a result, such a constraint incurs a sparsity
degradation. Fig. \ref{SD} shows the sparsity degree of the models with the
most sparse hidden state for different batch sizes while incurring no accuracy
degradation (i.e., the sweet spots). The run-time energy efficiency and performance of the proposed
accelerator for both dense and sparse models and different batch sizes are
reported in Fig. \ref{performance} and Fig. \ref{energy}, respectively.
Performing the recurrent computations using the sparse hidden state results in
up to 5.2$\times$ speedup and energy efficiency compared to the most
energy-efficient dense model.

\section{Related Work}
During the past few years, many works have focused on designing custom
accelerators for CNNs. For instance, DianNao exploits straight-forward
parallelism by adopting an array of multiply-accumulate units to accelerate the
computations of CNNs \cite{diannao}. However, such an approach requires
numerous memory accesses to the off-chip memory, which dominates its power
consumption. DaDianNao was introduced in \cite{dadiannao} to eliminate the
memory accesses to the off-chip memory by storing the weights on-chip. Cnvlutin
\cite{cnvlutin} relies on DaDianNao architecture and exploits the intrinsic
sparsity among activations incurred by using the ReLU as the non-linear
function to speedup the convolutional process. Cnvlutin showed that skipping
the ineffectual computations with zero-valued activations can improve the
performance by up to 1.55$\times$. In \cite{EIE}, a custom accelerator was
proposed to perform vector-matrix multiplications of fully-connected networks
using sparse and compressed weights. While all the aforementioned architectures
have proven to be effective, they accelerate only either CNNs or
fully-connected networks, not LSTMs \cite{ese}.\par

In \cite{ese}, a custom architecture, called ESE, was introduced to accelerate
the computations of LSTMs. ESE exploits the sparsity among the weight matrices
to speed up the recurrent computations. More precisely, it was shown that
performing the recurrent computations on the model with sparse weights is
4.2$\times$ faster than the model with dense weights when running both models
on ESE. In \cite{lstmarc}, a new sparse matrix format and its accelerator
called CBSR were introduced to improve over ESE architecture. This work showed
that using the new sparse format improves the performance by
25\%$\sim$30\% over ESE. The two above works exploited the sparsity among the
weights in LSTMs to speed up the recurrent computations. However, our work in
this paper takes a completely different approach by first learning the ineffectual
information in the hidden state and then exploiting the sparsity among them to
accelerate the recurrent computations while using dense weights. Fig.
\ref{energy_comp} compares the proposed accelerator in this paper with ESE and
CBSR. In fact, our work outperforms both ESE and CBSR in terms of performance
by factors of 1.9$\times$  and 1.5$\times$ respectively. It is worth mentioning
that we have used the improvement factor of CBSR over ESE to estimate the
performance of CBSR architecture in Fig. \ref{energy_comp}. Moreover, ESE
yields a peak energy efficiency of 61.5 GOPS/W on a Xilinx FPGA while our
accelerator results in a peak energy efficiency of 4.8 TOPS/W on an ASIC
platform. Therefore, a direct comparison in terms of energy efficiency does not
construct a fair comparison.

\begin{figure}[!t]
\centering
\small
\scalebox{0.8}{
\begin{tikzpicture}
  \centering
  \begin{axis}[
        ybar, axis on top,
        height=4cm, width=6cm,
        bar width=0.4cm,
        ymajorgrids, tick align=inside,
        major grid style={draw =white},
        enlarge y limits={value=.1,upper},
        ymin=0, ymax=5,
        axis x line*=bottom,
        axis y line*=left,
        y axis line style={opacity=0},
        tickwidth=0pt,
        enlarge x limits={abs = 1.5cm},
        x = 2cm,
        legend style={
            at={(0.5,-0.2)},
            anchor=north,
            legend columns=-1,
            /tikz/every even column/.append style={column sep=0.5cm}
        },
        ylabel style={align=center},
        ylabel={Performance\\(TOPS)},
        symbolic x coords={
         This work, ESE, CBSR  },
       xtick=data,
       nodes near coords=\rotatebox{90}{
        \pgfmathprintnumber[precision=1]{\pgfplotspointmeta}
       }
    ]
    \addplot [draw=none, fill=red!40] coordinates {
      (This work, 4.8)
      (ESE, 2.5)
      (CBSR,3.3) };

  \end{axis}
  \end{tikzpicture}}
\caption{Peak performance of the state-of-the-art accelerators.}
\label{energy_comp}
\vspace{-5pt}
\end{figure}
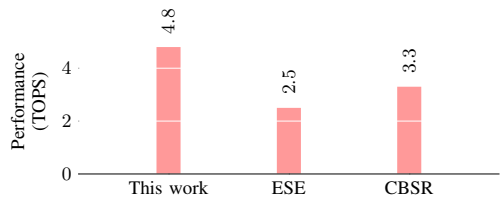

\section{Conclusion}
In this paper, we first introduced a new training scheme that learns to retain
only the important information in the hidden state $h_{t-1}$ of LSTMs. We then
showed that the proposed method can prune away over 90\% of the hidden state
values without incurring any degradation on the prediction accuracy over a set
of temporal tasks. We also introduced a new accelerator that performs the
recurrent computations on both dense and sparse representations of the hidden
state vector. Finally, we showed that performing the recurrent computations on
the sparse models results in up to 5.2$\times$ speedup and energy efficiency
over their dense counterparts.


\bibliographystyle{IEEEtran}
\bibliography{Bibliography}

\end{document}